\pdfoutput=1

\documentclass[11pt]{article}

\usepackage{ACL2023}
\usepackage{times}
\usepackage{latexsym}

\usepackage[T1]{fontenc}

\usepackage[utf8]{inputenc}

\usepackage{microtype}

\usepackage{graphicx}
\usepackage{multicol}
\usepackage{multirow}
\usepackage{amsmath}
\usepackage{amssymb}
\usepackage{bbding}
\usepackage{subfigure}
\usepackage{colortbl}
\usepackage{pifont}
\usepackage{bbm}
\usepackage{makecell}
\usepackage{bbding}
\usepackage{mathrsfs}
\usepackage{bm}
\usepackage{CJKutf8}
\usepackage{booktabs}
\usepackage{tabu}
\usepackage{fontawesome}

\usepackage[linesnumbered,ruled,vlined]{algorithm2e}

\newcommand{\instruct}{\textsc{x-Instruct}}
\newcommand{\ourmethod}{Qwen2.5-xCoder}

\title{Multi-Agent Collaboration for Multilingual Code Instruction Tuning}

\author{
  Jian Yang\textsuperscript{\rm 1}, 
  {\bf Wei Zhang}\textsuperscript{\rm 1},
  Jiaxi Yang\textsuperscript{\rm 1}, 
  {\bf Yibo Miao}\textsuperscript{\rm 2}, 
  {\bf Shanghaoran Quan}\textsuperscript{\rm 1}, 
  {\bf Zhenhe Wu}, 
  \\
  {\bf Qiyao Peng},
  {\bf Liqun Yang}, 
  {\bf Tianyu Liu}\textsuperscript{\rm 1},
  {\bf Zeyu Cui}\textsuperscript{\rm 1}, 
  {\bf Binyuan Hui}\textsuperscript{\rm 1}, 
  {\bf Junyang Lin}\textsuperscript{\rm 1} \\
   \textsuperscript{\rm 1}Alibaba Group;  
   \textsuperscript{\rm 2}Shanghai Jiao Tong University \\
   \textsuperscript{\rm 3}Shenzhen Institutes of Advanced Technology, Chinese Academy of Sciences;\\
  \textsuperscript{\rm 4}University of Chinese Academy of Sciences; \\
   \texttt{\{yj411294\}@alibaba-inc.com} \\
}

\begin{document}
\begin{CJK*}{UTF8}{gbsn}
\maketitle
\begin{abstract}
Recent advancement in code understanding and generation demonstrates that code LLMs fine-tuned on a high-quality instruction dataset can gain powerful capabilities to address wide-ranging code-related tasks. However, most previous existing methods mainly view each programming language in isolation and ignore the knowledge transfer among different programming languages. To bridge the gap among different programming languages, we introduce a novel multi-agent collaboration framework to enhance multilingual instruction tuning for code LLMs, where multiple language-specific intelligent agent components with generation memory work together to transfer knowledge from one language to another efficiently and effectively. Specifically, we first generate the language-specific instruction data from the code snippets and then provide the generated data as the seed data for language-specific agents. Multiple language-specific agents discuss and collaborate to formulate a new instruction and its corresponding solution (A new programming language or existing programming language), To further encourage the cross-lingual transfer, each agent stores its generation history as memory and then summarizes its merits and faults. Finally, the high-quality multilingual instruction data is used to encourage knowledge transfer among different programming languages to train \ourmethod{}.  Experimental results on multilingual programming benchmarks demonstrate the superior performance of \ourmethod{} in sharing common knowledge, highlighting its potential to reduce the cross-lingual gap.
\end{abstract}

\section{Introduction}
Recent advancements \cite{gpt4,phi_1,phi_1.5,code_llama,starcoder2,qwen25coder} in code understanding and synthesis have seen a transformative shift from small machine learning or deep learning models toward large language models (LLMs) based on the Transformer architecture. The emergence of code LLMs equipped with instruction tuning has advanced a revolutionary step in many code downstream tasks, where LLMs are first trained on massive codebases with autoregressive objectives and then aligned to human preferences and downstream tasks. Code LLMs can understand complex programming problems and produce code closely mirroring user intents. 

\begin{figure}[t]
\centering
\includegraphics[width=1.0\columnwidth]{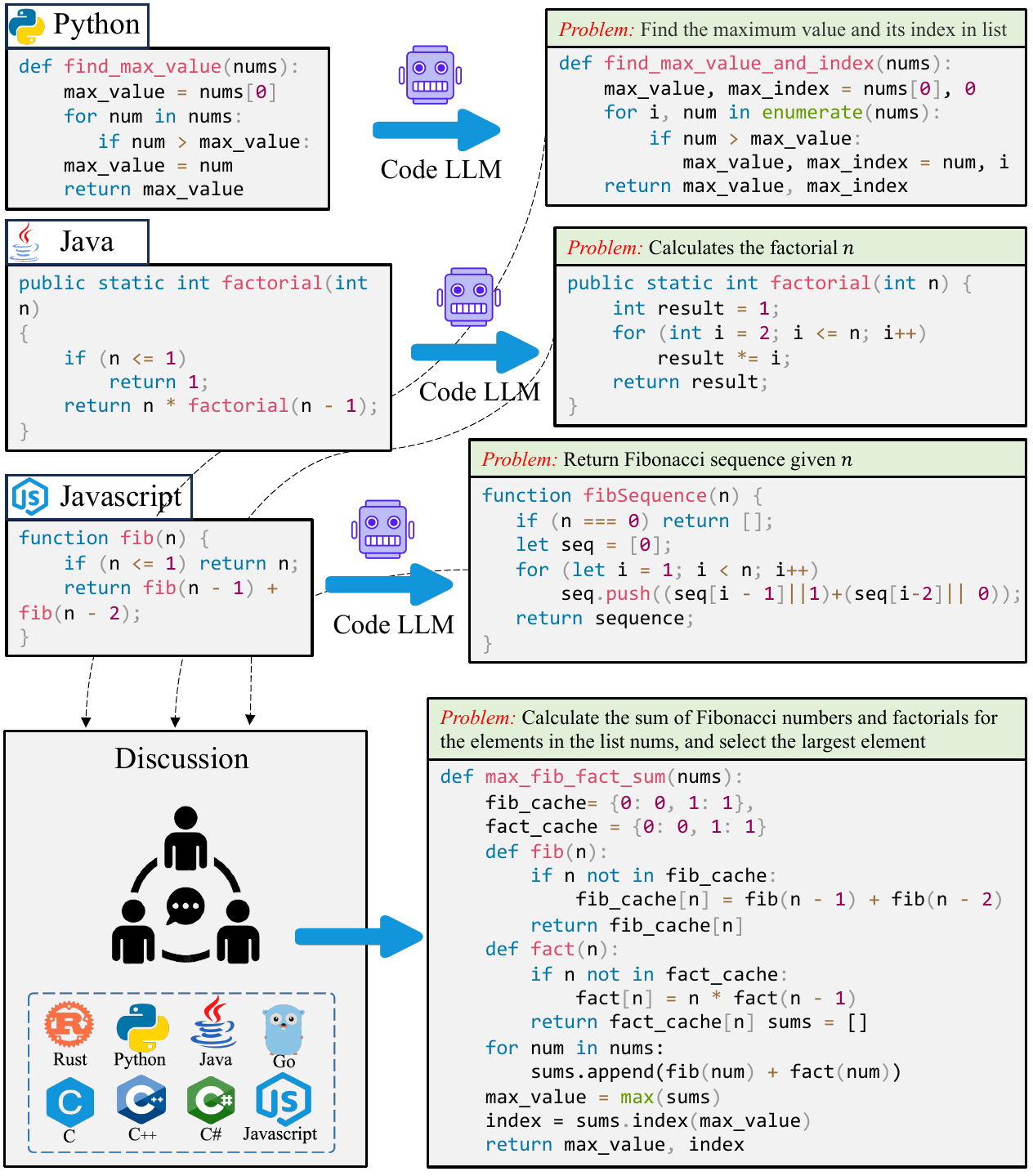}
\caption{An example of \ourmethod{}. The Code LLM solves the code generation question by ``translating'' the pseudocode description (Universal Code) into executable code of the target programming language.}
\vspace{-10pt}
\label{fig:intro}
\end{figure}

In the landscape of AI-driven code-related tasks, proprietary models such as ChatGPT and GPT-4 have gained dominance. The open-source community is making strides to narrow this gap, where Self-Instruct \cite{self_instructions} enhances the instruction-following capabilities of open-source LLMs. Code Alpaca \cite{codealpaca} uses ChatGPT to synthesize instructions with Self-Instruct. Further, Evol-Instruct evolves Code Alpaca by attempting to make code instructions more complex and fine-tune the code LLM  with the evolved data. A series of instruction data construction methods are proposed to generate diverse, high-quality instruction data from code snippets. However, these methods mainly focus on each programming language in isolation, ignoring the knowledge transfer among different programming languages.

To minimize the gap among different programming languages, we propose a novel multi-agent collaboration framework
to generate the high-quality instruction dataset \instruct{} of multilingual programming languages, which is used to fine-tune our proposed model \ourmethod{}. Specifically, we employ a multi-agent system where each agent is specialized in a different programming language to facilitate efficient and effective knowledge transfer across languages. Initially, we generate language-specific instruction data from code snippets, with each sample serving as the basis for an individual language-specific agent. These agents then engage in a collaborative discussion to synthesize new instructions, applicable to either a new or an existing programming language, along with their corresponding solutions. To enhance cross-lingual learning, each agent retains a record of its generation history, using this memory to assess its strengths and weaknesses. This iterative process results in high-quality, multilingual instruction data that is instrumental in training our method and fostering knowledge exchange among diverse programming languages.

\ourmethod{} is evaluated on the Python benchmark, including HumanEval~\cite{humaneval} and MBPP~\cite{mbpp}, and the extended multilingual benchmark MultiPL-E, comprised of Python, Java, CPP, C-sharp, Typescript, PHP, and Bash. The Experimental results demonstrate that \ourmethod{} consistently outperforms the previous baselines. Empirical studies show that \ourmethod{} can effectively transfer knowledge of data in different languages to each other and thus help to alleviate the
negative language interference among various languages. 
The contributions are summarized as follows:
\begin{itemize}
    \item We introduce a multilingual multi-agent framework, where multiple agents participate in a collaborative discussion to synthesize new instructions and the corresponding answers. These cooperative agents work together towards a shared goal, typically exchanging information to enhance a collective solution.
    \item Based on the code snippets extracted from the open-source code, we leverage the multilingual multi-agent framework to create a multilingual programming instruction dataset \instruct{} to improve the cross-lingual capabilities of the code LLM.
    \item To validate the effectiveness of our method,  we introduce a series of \ourmethod{} models fine-tuned on our data generation strategy based on Code Llama, and Deepseek-Coder.
\end{itemize}

\begin{figure*}[t]
\begin{center}
    \includegraphics[width=1.0\textwidth]{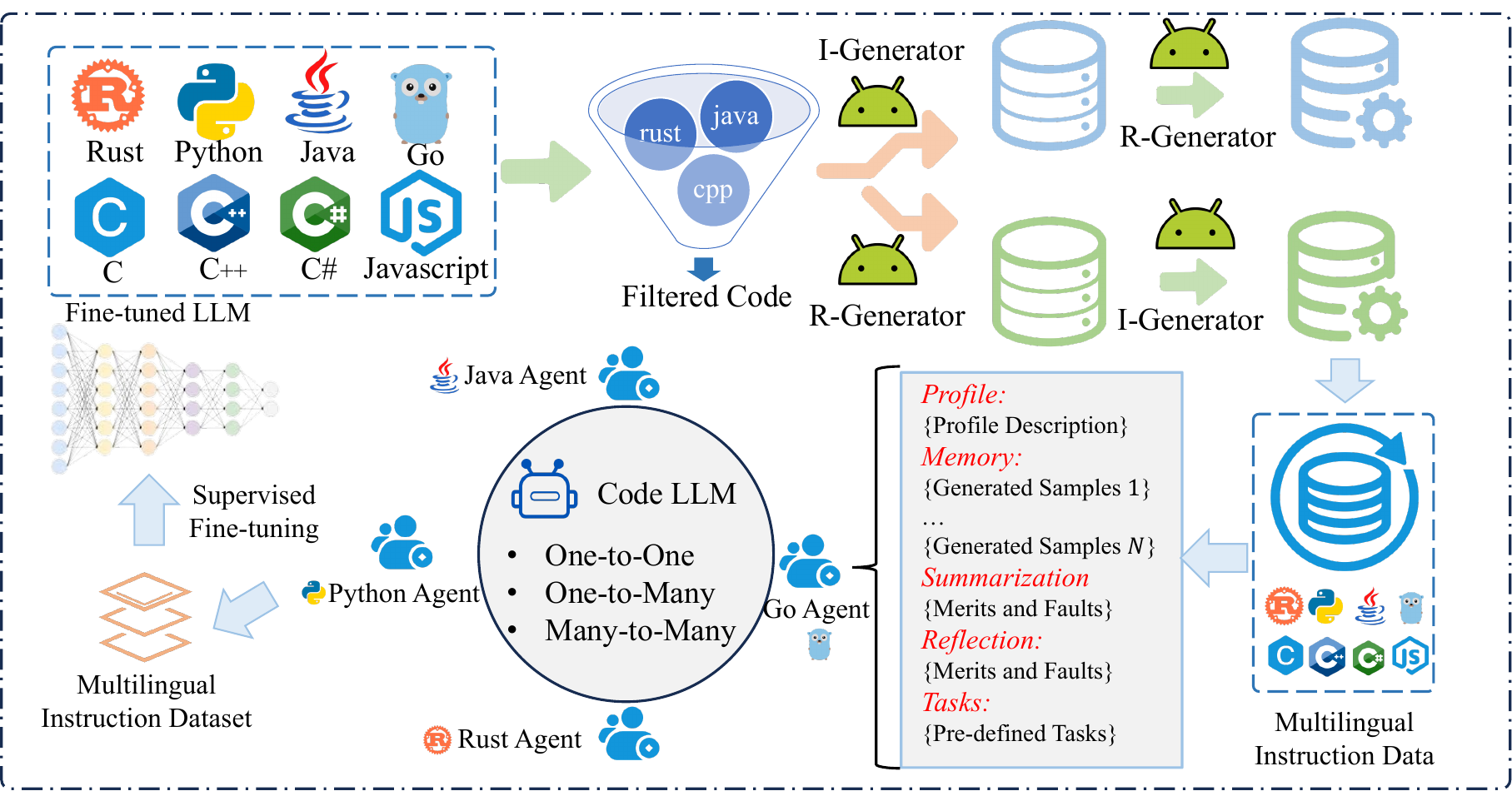}
    \caption{Overview of multilingual multi-agent data generation framework. we first construct the multilingual instruction dataset from the code snippets. We introduce a multi-agent framework, with each agent possessing expertise in a different programming language, allowing for efficient knowledge transfer across various languages. 
    ``R-Generator'' generates the responses based on the instruction while ``I-Generator'' generates the instruction based on the responses.
    Each snippet is assigned to a language-specific agent who uses it to create individual instructions. The agents then collaborate, using their specialized knowledge to create new instructions that can be applied to either a new or existing programming language, along with the appropriate solutions. To improve cross-lingual learning, agents maintain a history of their generated instructions, allowing them to identify their strengths and areas for improvement. Through this collaborative process, we produce high-quality multilingual instruction data for instruction tuning.}
    \label{fig:model}
    \vspace{-10pt}
\end{center}
\end{figure*}
\section{\ourmethod{}}
\subsection{Model Overview}
In Figure \ref{fig:model}, we develop a multi-agent framework to construct a multilingual instruction dataset from code snippets. Each agent in the framework specializes in a different programming language, facilitating effective knowledge transfer between languages. The code snippets are assigned to the respective language-specific agents, who then generate individual instructions. These agents collaborate, leveraging their expertise to create new instructions that can be applied to various programming languages, along with corresponding solutions. To enhance cross-lingual learning, agents keep a record of their generated instructions to identify their strengths and areas for development. This collaborative approach allows us to produce a high-quality multilingual instruction dataset that can be used for instruction tuning.

\begin{figure}[h!]
\centering
\includegraphics[width=0.85\columnwidth]{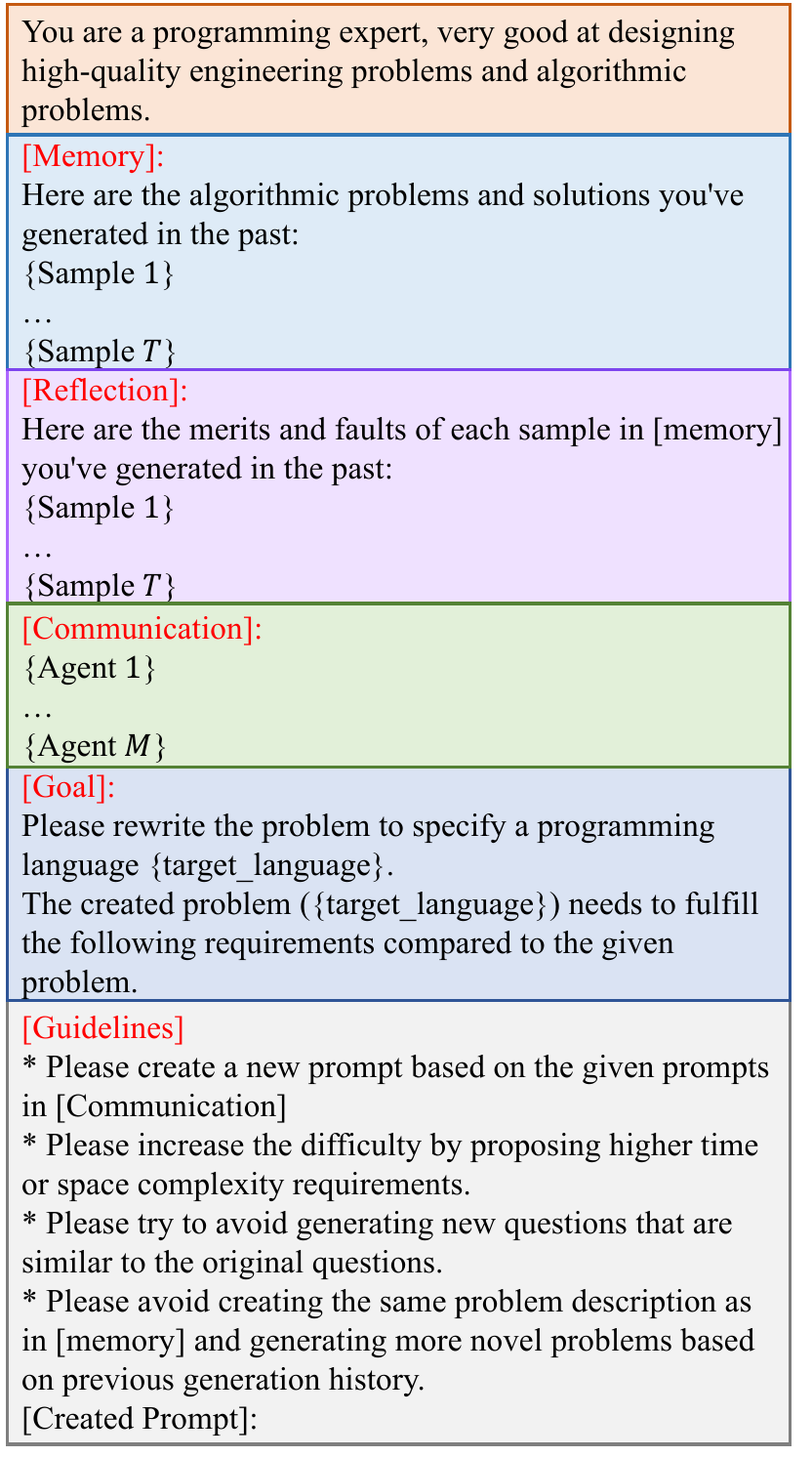}
\caption{Prompt of the multilingual multi-agent framework.}
\label{fig:multi_agent_prompt}
\end{figure}

\begin{figure}[h!]
\centering
\includegraphics[width=0.85\columnwidth]{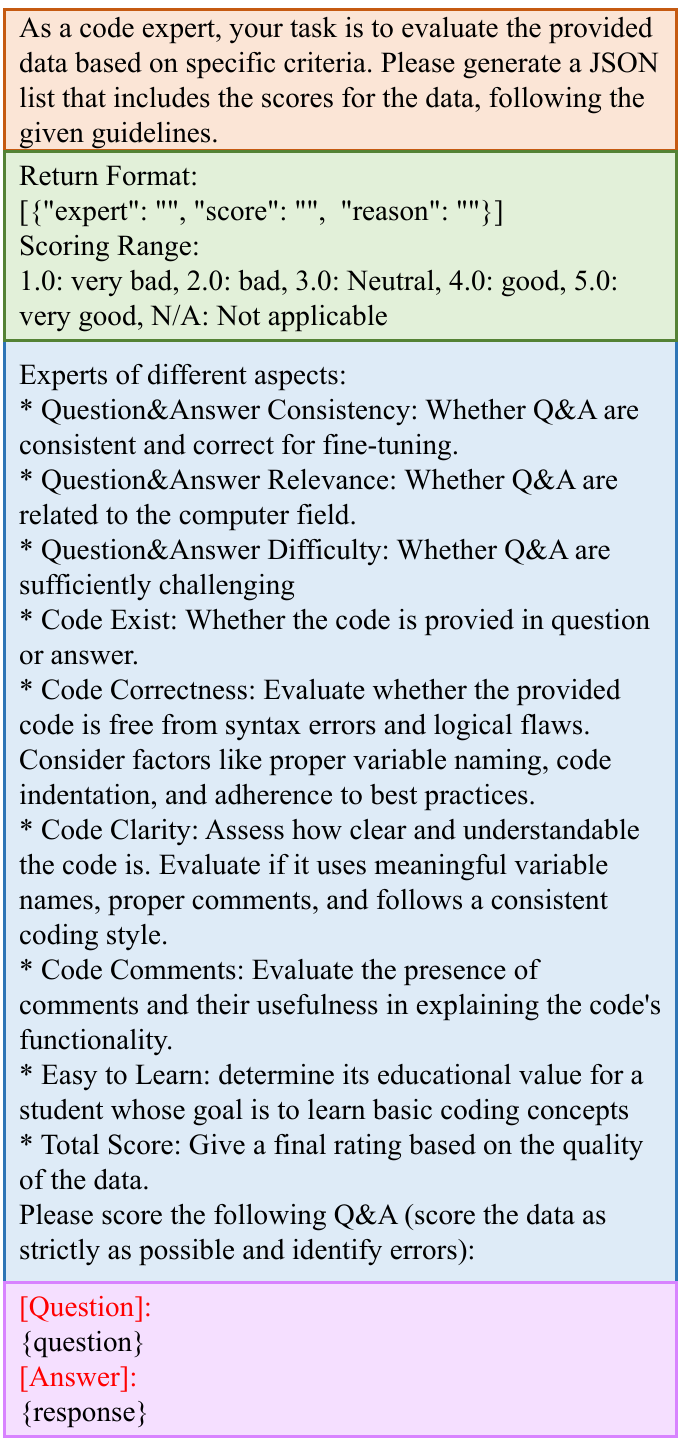}
\caption{Prompt of evaluation.}
\vspace{-10pt}
\label{fig:evaluation_prompt}
\end{figure}

\subsection{Seed Instruction Dataset}
\paragraph{Instruction from Code Snippet.} For the unsupervised data (code snippets) massively existing in many websites (e.g. GitHub), we try to construct the supervised instruction dataset. Specifically, we use the LLM to generate the instruction $q$ from the code snippets within 1024 tokens and then we use the code LLM to generate the response $a$. Finally, we use the LLM scorer in Figure \ref{fig:evaluation_prompt} to filter the low-quality ones to obtain the final pair $(q,a)$. Given the code snippets of different programming languages $L_{k} \in \{L_{k}\}_{k=1}^{K}$, we construct instruction dataset $D_{s_{1}}=\{D_{s_{1}}^{L_{k}}\}_{k=1}^{K}$ from the code snippets. ($K$ is the number of programming languages)
\paragraph{Response from Code Snippet.} To increase the diversity of the instruction dataset. Conversely, we first generate the responses from the code and then prompt the LLM to generate instructions. Then we use the LLM scorer to filter the low-quality to obtain the final pair $(q,a)$. Similarly, given the code snippets of different programming languages $L_{k} \in \{L_{k}\}_{k=1}^{K}$, we can construct instruction dataset $D_{s_{2}}=\{D_{s_{2}}^{L_{k}}\}_{k=1}^{K}$ from the code snippets. To fully unleash the potential of our proposed method, we combine two parts instruction dataset $D_{s_{1}} \cup D_{s_{2}}$ as the seed data for multi-agent data generation framework.

\subsection{Language Agent} 
Figure \ref{fig:model} shows the overall framework of the multi-agent framework to generate the new samples.

\paragraph{Instance-level Agent.} Given the seed instruction dataset $D_{s}$, each instance $(q,a) \in D_{s}$ is used to initialize the agent $\mathcal{A} = \{p, m, r, O\}$, where $\mathcal{A}$ contains the pre-defined agent profile $p$ (task definition), the agent operation $o \in O=\{o_{1}\,\dots,o_{a}\}$, the memory of the generated history $m =\{m_{1},\dots,m_{T}\}$, the reflection $r = \{r_1,\dots, r_{c}\}$. $O$ contains different evolution operations, such as increasing difficulty and adding more reasoning steps. $m$ is comprised of $T$ history generated samples, where $m_{i}=(q_{i},a_{i}) \in m$ ($q_{i}$ and $a_{i}$ are the question and answer generated by the agent $\mathcal{A}$). 
\paragraph{Memory Initialization\&Update.} The memory module $m$ is essential for the abilities of an agent to gather, retain, and apply knowledge gained from interactions. 
Initially, the agent $\mathcal{A}$ does not generate any samples, and thus the memory is initialized by the seed data ($m = \{q_{s}, a_{s}\}$). Memory updating is the process of recording new information, thereby updating the knowledge base of the agent with fresh insights or observations. When the agent generates the new sample $(q,a)$, the pair is added into the memory $m$, where the similarity between $(q,a)$ and the samples already in $m$ is less than a certain threshold to prevent duplicates. $m$ is a priority queue with capacity size $T$. When the capacity is full, the sample with the lowest evaluated score is first removed from the queue.

\paragraph{Memory Relfection.}
The goal is to equip agents with the ability to autonomously generate summaries of their experiences and draw inferences that reach beyond simple data processing. After the agent $\mathcal{A}$ synthesizes one new sample $(q,a)$, we use the LLM scorer to evaluate the sample in many aspects and obtain the reflection $r$ in Figure \ref{fig:evaluation_prompt}.

\subsection{Cross-lingual Discussion}
The communication between agents in our proposed multi-agent framework is the critical infrastructure supporting collective intelligence. We introduce two types of communication structures, including centralized communication and distributed communication. 
When multiple agents try to synthesize new samples, each agent will randomly select a sample from its memory for further data synthesis. (for the first data synthesis, the agents uses the provided seed data $D_{s_{1}} \cup D{s_{2}}$ to synthesize new samples.)

\paragraph{Centralized Discussion.}
Centralized communication involves a central agent coordinating the communication, with other agents primarily interacting through this central node. Given $d$ agents \{$\mathcal{A}_{1},\dots,\mathcal{A}_{d}$\}($\mathcal{A}_{1}$ is the main agent and others are auxiliary agents), the prompt $\mathcal{F}$ is used to generate the new sample $(q,a)$ mainly based on $\mathcal{A}_{1}$, where the process can be described as $\mathcal{F}(\mathcal{A}_{1};\mathcal{A}_{2},\dots,\mathcal{A}_{d})$.

\paragraph{Parallel Discussion.}
Parallel discussion equally regards all agents $\mathcal{A}_{1},\dots,\mathcal{A}_{d}$, the prompt is used to consider all agents and generate a new sample, where the process is described as
$\mathcal{F}(\mathcal{A}_{1},\mathcal{A}_{2},\dots,\mathcal{A}_{d})$.

\paragraph{Data Generation.} Based on the seed instruction dataset $D_{s}$, we adopt the multilingual agent framework to create the multilingual instruction dataset $D = \{D^{L_{k}}\}_{k=1}^{K}$ ($K$ is the number of languages). Finally, the generated multilingual data from different agents comprise the corpus $D_{s_{3}}$.

\subsection{Multilingual Code Instruction Tuning}
Given the multilingual corpora $D=\{D^{L_{k}}\}_{k=1}^{K}$, the training objective of the $\mathcal{L}_{\text{SFT}}$ can be described:
\begin{align}
    \mathcal{L}_{\text{SFT}} = -\sum_{k=1}^{K} \mathbb{E}_{q,a \sim D_{L_{k}}} \left[ \log P(a|q; \mathcal{M}) \right]
    \label{code_sft}
\end{align}where $q$ and $a$ are the question and answer pair. $\mathcal{L}_{\text{SFT}}$ is the instruction fine-tuning objective.

\subsection{Multilingual Code DPO}
\begin{SmallEquation}
\begin{align}
\mathcal{L}_{\text{DPO}} = -\mathbb{E}_{(x,y^{+},y^{-})\sim \mathcal{D}} \left[\log \sigma(\beta(\frac{\pi_{\theta}(y^{+}|x)}{\pi_{\text{r}}(y^{+}|x)} - \frac{\pi_{\theta}(y^{-}|x)}{\pi_{\text{r}}(y^{-}|x)}) \right]
\end{align}
\end{SmallEquation}where $(y^{+}, y^{-})$ is the positive and negative pair. $\sigma$ denotes the sigmoid function. After generating the SFT data, we further leverage the multi-agent collaboration data generation framework to synthesize the DPO data. For the sample query, we prompt the SFT model to sample 128 responses and use the code execution to verify the correctness of the generated code snippet. We feed generated test cases from LLM into the code snippet to verify the correctness of the function under the multilingual environment. The DPO data is denoted as the $D_{s_{4}}$. Finally, we use $D_{s_{1}} \cup D_{s_{2}} \cup D_{s_{3}}$ for supervised fine-tuning and $D_{s_{4}}$ for preference learning.

\section{Experiments}
\subsection{Implementation Details}
For the code snippets collected from GitHub, we choose nearly 100K code snippets from different languages (Python, Java, CPP, C-sharp, Typescript, PHP, and Bash) to construct the synthetic instruction dataset. Finally, we obtain the instruction dataset \instruct{} containing nearly 95K training samples.
We utilize Qwen2.5-Coder as the foundational code LLMs for supervised fine-tuning. We fine-tune these foundation LLMs on \instruct{}. \ourmethod{} (32B) is fine-tuned on the 128 NVIDIA H800-80GB GPUs\footnote{\url{https://github.com/QwenLM/Qwen2.5-Coder/tree/main/finetuning/sft}}. The learning rate first increases into 5e-5 with 100 warmup steps and then adopts a cosine decay scheduler. We adopt the Adam optimizer with a global batch size of 1024 samples, truncating sentences to 2048 tokens. 

\subsection{Evaluation Metrics}
\paragraph{Pass@k.} We adopt the Pass@k metric \cite{codegeex} for evaluation by using test cases to verify the correctness of the generated code. For the fair comparison, we report the greedy Pass@1 scores for all LLMs in our work.

\subsection{Instruction Dataset}
We use the created instruction dataset we combine three parts instruction dataset $D_{s_{1}} \cup D_{s_{2}} \cup D_{s_{3}}$ for supervised fine-tuning, comprising nearly 97.3K instruction samples. The created samples $D_{s_{1}} \cup D_{s_{2}}$ from code snippets contain 30K samples. The existing instruction dataset $D_{s_{3}}$ contains nearly 67K samples. The data used for DPO is comprised of 133K samples.

\subsection{Baselines}
\paragraph{Proprietary Models.}
GPT-3.5 and GPT-4~\cite{gpt4} are both LLMs developed by OpenAI based on a neural architecture known as generative pre-trained Transformers (GPT)~\cite{gpt}. They are both trained on massive datasets of text and code, allowing them to generate human-quality text, translate languages, and write different kinds of creative content. GPT-3.5 Turbo (i.e., ChatGPT) and GPT4 achieve excellent performance in various code understanding and generation tasks.

\paragraph{Open-Source Models.}
To narrow the gap between open-source and closed-source models, a series of open-source models and instruction datasets are proposed to improve the code foundation LLMs and bootstrap the instruction-following ability of code LLMs. We compare the following code LLMs: Qwen-Coder~\cite{qwen25coder}, Code Llama~\cite{code_llama}, and DeepSeek-Coder~\cite{deepseek_coder} with different model sizes are introduced into the based model. 

\subsection{Evaluation Benchmark}
\paragraph{EvalPlus.} 
The HumanEval test set~\cite{humaneval} is a crafted collection of 164 Python programming problems to test the abilities of code generation models. For each problem, there are roughly 9.6 test cases to check whether the generated code works as intended. Humaneval has become the most popular benchmark to measure how well these code-writing AI models perform, making it a key tool in the field of AI and machine learning for coding.
The MBPP dataset~\cite{mbpp}, comprising approximately 1,000 Python programming challenges sourced from a crowd of contributors, is tailored for beginners in programming, focusing on core principles and the usage of the standard library. The MBPP test set comprised of 500 problems is selected to evaluate the few-shot inference of the code LLMs. 

\paragraph{MultiPL-E.}
The MultiPL-E test set~\cite{multipl_e} translates the original HumanEval test set to other languages and we report the scores of the languages Python, Java, CPP, Typescript, Javascript, PHP, and Bash.

\subsection{Baselines}

\paragraph{Proprietary Models.}
GPT-3.5 and GPT-4~\cite{gpt4} are both LLMs developed by OpenAI based on a neural architecture known as generative pre-trained Transformers (GPT). They are both trained on massive datasets of text and code, allowing them to generate human-quality text, translate languages, and write different kinds of creative content. GPT-3.5 Turbo (i.e., ChatGPT) and GPT4 achieve excellent performance in various code understanding and generation tasks.

\paragraph{Open-Source Models.}
To narrow the gap between open-source and closed-source models, a series of open-source models and instruction datasets are proposed to improve the code foundation LLMs and bootstrap the instruction-following ability of code LLMs. Starcoder~\cite{starcoder}, Code Llama~\cite{code_llama}, and DeepSeek-Coder~\cite{deepseek_coder} with different model sizes are introduced into the based model. OctoCoder, WiazrdCoder, MagiCoder, WaveCoder

\section{Base Models}
\paragraph{Qwen2.5-Coder.} The Qwen2.5-Coder~\cite{deepseek_coder} series contains a range of open-source code models with sizes from 0.5B to 32B, pre-trained on 5.5 trillion tokens from the Qwen2.5 base model. These models are fine-tuned on a high-quality code corpus, using a 32K token window for a fill-in-the-middle task to improve code generation and completion.

\begin{table*}[h!]
 \centering
 \resizebox{0.85\textwidth}{!}{
 \begin{tabular}{lr|cccc|cccccccc|c}
 \toprule
 \textbf{Model} & \textbf{Size} & HE & HE+ & MBPP & MBPP+ & Python & Java & C++ & C\# & TS & JS & PHP & Bash & \textbf{Avg.} \\
 \midrule
 \multicolumn{15}{c}{\textbf{Closed-APIs}} \\ \midrule
 Claude-3.5-Sonnet-20240620 & \faLock{} & 89.0 & 81.1 & 87.6 & 72.0 & 89.6 & 86.1 & 82.6 & 85.4 & 84.3 & 84.5 & 80.7 & 48.1 & 80.2 \\
 Claude-3.5-Sonnet-20241022 & \faLock{} & 92.1 & 86.0 & 91.0 & 74.6 & 93.9 & 86.7 & 88.2 & \underline{87.3} & 88.1 & 91.3 & 82.6 & 52.5 & 83.8 \\
 GPT-4o-mini-2024-07-18 & \faLock{} & 87.8 & 84.8 & 86.0 & 72.2 & 87.2 & 75.9 & 77.6 & 79.7 & 79.2 & 81.4 & 75.2 & 43.7 & 75.0 \\
 GPT-4o-2024-08-06 & \faLock{} & 92.1 & 86.0 & 86.8 & 72.5 & 90.9 & 83.5 & 76.4 & 81.0 & 83.6 & 90.1 & 78.9 & 48.1 & 79.1 \\
 o1-mini & \faLock{} & \underline{97.6} & \underline{90.2} & \underline{93.9} & \underline{78.3} & 95.7 & \underline{90.5} & \underline{93.8} & 77.2 & \underline{91.2} & 92.5 & 84.5 & \underline{55.1} & 85.1 \\
 o1-preview & \faLock{} & 95.1 & 88.4 & 93.4 & 77.8 & \underline{96.3} & 88.0 & 91.9 & 84.2 & 90.6 & \underline{93.8} & \underline{90.1} & 47.5 & \underline{85.3} \\
 \midrule
 \multicolumn{15}{c}{\textbf{0.5B+ Models}} \\ \midrule
 Qwen2.5-Coder-0.5B-Instruct & 0.5B & 61.6 & 57.3 & 52.4 & 43.7 & 61.6 & 57.3 & 52.4 & 43.7 & 50.3 & 50.3 & 52.8 & 27.8 & 49.6 \\
  \rowcolor{gray!15} \ourmethod{} (SFT) & 0.5B & 69.5 & 66.5 & \underline{52.6} & \underline{45.5} & 68.9 & 57.6 & 53.4 & 67.7 & 63.5 & \underline{66.5} & \underline{57.8} & 33.5 & 57.1 \\
 \rowcolor{gray!15} \ourmethod{} (DPO) & 0.5B & \underline{72.6} & \underline{67.1} & 51.9 & 45.2 & \underline{72.0} & \underline{58.2} & \underline{54.0} & \underline{68.4} & \underline{66.7} & 64.0 & \underline{57.8} & \underline{36.7} & \underline{58.0} \\
 \midrule
 \multicolumn{11}{c}{\textbf{1B+ Models}} \\ \midrule
 DS-Coder-1.3B-Instruct & 1.3B & 65.9 & 60.4 & 65.3 & 54.8 & 65.2 & 51.9 & 45.3 & 55.1 & 59.7 & 52.2 & 45.3 & 12.7 & 48.4 \\
 Yi-Coder-1.5B-Chat & 1.5B & 69.5 & 64.0 & 65.9 & 57.7 & 67.7 & 51.9 & 49.1 & 57.6 & 57.9 & 59.6 & 52.2 & 19.0 & 51.9 \\
 Qwen2.5-Coder-1.5B-Instruct & 1.5B & 70.7 & 66.5 & 69.2 & 59.4 & 71.2 & 55.7 & 50.9 & 64.6 & 61.0 & 62.1 & 59.0 & 29.1 & 56.7 \\
 \rowcolor{gray!15} \ourmethod{} (SFT) & 1.5B & \underline{85.4} & \underline{79.9} & 69.3 & 59.5 & 70.1 & 68.4 & 66.5 & 68.4 & 71.1 & 72.7 & \underline{70.2} & 44.3 & 65.9 \\
 \rowcolor{gray!15} \ourmethod{} (DPO) & 1.5B & 65.2 & 60.4 & \underline{70.9} & \underline{61.6} & \underline{74.4} & \underline{69.6} & \underline{67.7} & \underline{70.9} & \underline{75.5} & \underline{74.5} & 69.6 & \underline{44.9} & \underline{67.5} \\
 \midrule
 \multicolumn{15}{c}{\textbf{3B+ Models}} \\ \midrule
 Qwen2.5-Coder-3B-Instruct & 3B & 84.1 & \underline{80.5} & 73.6 & 62.4 & \underline{83.5} & \underline{74.7} & 68.3 & 78.5 & 79.9 & 75.2 & 73.3 & 43.0 & \underline{72.1} \\
  \rowcolor{gray!15} \ourmethod{} (SFT) & 3B & 84.8 & 79.9 & 69.3 & 61.1 & 67.7 & 57.6 & 39.8 & 28.5 & 43.4 & 55.3 & 38.5 & 31.6 & 42.1 \\
 \rowcolor{gray!15} \ourmethod{} (DPO) & 3B & \underline{85.4} & \underline{80.5} & \underline{75.7} & \underline{66.7} & 80.5 & 11.4 & \underline{70.2} & \underline{80.4} & \underline{82.4} & \underline{82.6} & \underline{74.5} & \underline{43.7} & 63.6 \\
 \midrule
 \multicolumn{15}{c}{\textbf{6B+ Models}} \\ \midrule
 CodeLlama-7B-Instruct & 7B & 40.9 & 33.5 & 54.0 & 44.4 & 34.8 & 30.4 & 31.1 & 21.6 & 32.7 & - & 28.6 & 10.1 & - \\
 DS-Coder-6.7B-Instruct & 6.7B & 74.4 & 71.3 & 74.9 & 65.6 & 78.6 & 68.4 & 63.4 & 72.8 & 67.2 & 72.7 & 68.9 & 36.7 & 66.1 \\
 CodeQwen1.5-7B-Chat & 7B & 83.5 & 78.7 & 77.7 & 67.2 & 84.1 & 73.4 & 74.5 & 77.8 & 71.7 & 75.2 & 70.8 & 39.2 & 70.8 \\
 Yi-Coder-9B-Chat & 9B & 82.3 & 74.4 & 82.0 & 69.0 & 85.4 & 76.0 & 67.7 & 76.6 & 72.3 & 78.9 & 72.1 & 45.6 & 71.8 \\
 DS-Coder-V2-Lite-Instruct & 2.4/16B & 81.1 & 75.6 & 82.8 & 70.4 & 81.1 & \underline{76.6} & 75.8 & 76.6 & 80.5 & 77.6 & 74.5 & 43.0 & 73.2 \\
 Qwen2.5-Coder-7B-Instruct & 7B & \underline{88.4} & \underline{84.1} & \underline{83.5} & \underline{71.7} & \underline{87.8} & 76.5 & 75.6 & 80.3 & \underline{81.8} & \underline{83.2} & \underline{78.3} & \underline{48.7} & \underline{76.5} \\
 OpenCoder-8B-Instruct & 8B & 83.5 & 78.7 & 79.1 & 69.0 & 83.5 & 72.2 & 61.5 & 75.9 & 78.0 & 79.5 & 73.3 & 44.3 & 71.0 \\
 \rowcolor{gray!15} \ourmethod{} (SFT) & 7B & 86.0 & 79.9 & 81.2 & 67.2 & 84.1 & 39.2 & \underline{78.9} & 79.7 & \underline{81.8} & 78.9 & 68.3 & 46.2 & 67.6 \\
 \rowcolor{gray!15} \ourmethod{} (DPO) & 7B & 86.0 & 79.9 & 82.3 & 69.3 & 82.3 & 23.4 & 77.0 & \underline{81.6} & 81.1 & 82.0 & 65.8 & 47.5 & 65.5 \\
 \midrule
 \multicolumn{15}{c}{\textbf{13B+ Models}} \\ \midrule
 CodeLlama-13B-Instruct & 13B & 40.2 & 32.3 & 60.3 & 51.1 & 42.7 & 40.5 & 42.2 & 24.0 & 39.0 & - & 32.3 & 13.9 & - \\
 Starcoder2-15B-Instruct-v0.1 & 15B & 67.7 & 60.4 & 78.0 & 65.1 & 68.9 & 53.8 & 50.9 & 62.7 & 57.9 & 59.6 & 53.4 & 24.7 & 54.0 \\
 Qwen2.5-Coder-14B-Instruct & 14B & \underline{89.6} & \underline{87.2} & \underline{86.2} & 72.8 & 89.0 & 79.7 & \underline{85.1} & \underline{84.2} & \underline{86.8} & 84.5 & \underline{80.1} & 47.5 & \underline{79.6} \\
  \rowcolor{gray!15}  \ourmethod{} (SFT) & 14B & 88.4 & 83.5 & 84.1 & \underline{73.3} & 89.0 & \underline{81.0} & 75.2 & 81.6 & 84.3 & \underline{85.1} & 75.2 & 46.8 & 75.6 \\
  \rowcolor{gray!15} \ourmethod{} (DPO) & 14B & 89.0 & 84.8 & 83.1 & 72.0 & \underline{91.5} & \underline{81.0} & 78.9 & 82.9 & 85.5 & \underline{85.1} & 77.0 & \underline{48.1} & 76.9 \\
 \midrule
 \multicolumn{15}{c}{\textbf{20B+ Models}} \\ \midrule
 CodeLlama-34B-Instruct & 34B & 48.2 & 40.2 & 61.1 & 50.5 & 41.5 & 43.7 & 45.3 & 31.0 & 40.3 & - & 36.6 & 19.6 & - \\
 CodeStral-22B-v0.1 & 22B & 81.1 & 73.2 & 78.2 & 62.2 & 81.1 & 63.3 & 65.2 & 43.7 & 68.6 & - & 68.9 & 42.4 & - \\
 DS-Coder-33B-Instruct & 33B & 81.1 & 75.0 & 80.4 & 70.1 & 79.3 & 73.4 & 68.9 & 74.1 & 67.9 & 73.9 & 72.7 & 43.0 & 69.2 \\
 CodeLlama-70B-Instruct & 70B & 72.0 & 65.9 & 77.8 & 64.6 & 67.8 & 58.2 & 53.4 & 36.7 & 39.0 & - & 58.4 & 29.7 & - \\
 DS-Coder-V2-Instruct & 21/236B & 85.4 & 82.3 & 89.4 & 75.1 & 90.2 & \underline{82.3} & \underline{84.8} & 82.3 & 83.0 & 84.5 & \underline{79.5} & \underline{52.5} & \underline{79.9} \\
 Qwen2.5-Coder-32B-Instruct & 32B & \underline{92.7} & 87.2 & 90.2 & 75.1 & \underline{92.7} & 80.4 & 79.5 & \underline{82.9} & \underline{86.8} & \underline{85.7} & 78.9 & 48.1 & 79.4 \\
 Qwen2.5-32B-Instruct & 32B & 87.8 & 82.9 & 86.8 & 70.9 & 88.4 & 80.4 & 81.0 & 74.5 & 83.5 & 82.4 & 78.3 & 46.8 & 76.9 \\
 Qwen2.5-72B-Instruct & 32B & 85.4 & 79.3 & \underline{90.5} & \underline{77.0} & 82.9 & 81.0 & 80.7 & 81.6 & 81.1 & 82.0 & 77.0 & 48.7 & 75.1 \\ 
 Qwen2.5-SynCoder & 32B & \underline{92.7} & \underline{87.8} & 86.2 & 74.7 & 92.1 & 80.4 & 80.7 & 81.6 & 83.0 & \underline{85.7} & 77.6 & 49.4 & 78.8 \\ 
 \rowcolor{gray!15} \ourmethod{} (SFT) & 32B & 87.8 & 84.1 & 84.9 & 74.9 & 89.6 & 74.7 & 73.3 & 79.1 & 82.4 & 81.4 & 78.3 & 46.2 & 73.6 \\
 \rowcolor{gray!15} \ourmethod{} (DPO) & 32B & 89.0 & 86.0 & 85.4 & 74.9 & 90.9 & 76.6 & 72.7 & 79.1 & 83.6 & 81.4 & 78.9 & 49.4 & 74.5 \\
 \bottomrule
 \end{tabular}
 }
 \caption{The performance of different instruction LLMs on EvalPlus and MultiPL-E. ``HE'' denotes the HumanEval, ``HE+'' denotes the plus version with more test cases, and ``MBPP+'' denotes the plus version with more test cases.}
\label{tab:coding_multiple}
\end{table*}

\subsection{Main Results}
\label{subsec:results}
\paragraph{Python Code Generation.}
Table \ref{tab:coding_multiple} illustrates that \ourmethod{} significantly outperforms the base Code Llama and previous strong open-source baselines, minimizing the gap with GPT-3.5 and GPT-4. Particularly, \ourmethod{} outperforms the WizardCoder with 15B foundation LLM and Evol-Instruct techniques. Magicoder \cite{magicoder} and Wavecoder \cite{wavecoder} both prove the effectiveness of data construction from code snippets. 

\paragraph{Multilingual Code Generation.}
For the multilingual code generation task, our proposed model \ourmethod{} is evaluated on the MultiPL-E, including Python, Java, CPP, C-sharp, and other languages. The experimental results in Figure \ref{tab:coding_multiple} show that DS-Coder and Qwen2.5-Coder significantly outperform across all languages. The enhanced version, \ourmethod{}, further improves the multilingual performance of LLM, rivaling the CodeLlama-70B-Instruct model with only 14B parameters. Remarkably, we introduce the language-specific multi-agent framework to compose new samples for enhancing cross-lingual transfer among different programming languages.

\paragraph{Multilingual Code Understanding.}
Given the multilingual correct code snippet, the code LLM is tasked to generate an explanation of the code and then regenerate the code
only based on its own explanation. For the different backbones Code Llama and Deepseek-Coder, our method beats most previous methods, especially in other languages, which demonstrates that \instruct{} can bring multilingual agreement for different programming languages.

\section{Analysis}
\paragraph{Ablation Study.}
Given the synthetic instruction dataset, we can obtain the fine-tuned model \ourmethod{} based on the Qwen2.5-Coder-7B denoted as {\large \ding{172}}. Our multilingual instruction dataset contains four parts $\{D_{s_1},D_{s_2},D_{s_3},D_{s_4}\}$. $D_{s_1}$ is created from the code snippets by first generating instructions and then outputting the response while $D_{s_2}$ is developed from the code snippets by first generating responses and then synthesizing the instructions. $D_{s_3}$ is derived from our multilingual multi-agent framework. $D_{s_4}$ is derived from the temperature-based sampling for DPO training.
Specially, we can observe that {\large{\ding{174}}} drops a lot compared to {\large{\ding{173}}}. It indicates the significance of the dataset $D_{s_{3}}$ from the multi-agent data generation framework.
Table~\ref{ablation_study} summarizes the results of the ablation study of these datasets, which shows that our multilingual multi-agent framework can leverage code snippets and existing instruction datasets to transfer knowledge from Python to other languages.

\begin{table}[t]
\centering
\resizebox{0.97\columnwidth}{!}{
\begin{tabular}{c|c|ccccc}
\toprule
ID & Methods &  Python  & Java & C++ & C\# & Avg. \\
\midrule
{\large{\ding{172}}} & \ourmethod{} & \textbf{90.9} & \textbf{76.6} & \textbf{72.7} & \textbf{79.1} & \textbf{79.8} \\
{\large{\ding{173}}} & {\large{\ding{172}}} -  $D_{s_4}$             & 89.6 & 74.7 & 73.3 & 79.1  & 79.2 \\
{\large{\ding{174}}} & {\large{\ding{173}}} -  $D_{s_3}$             & 82.9 & 69.6 & 68.3 & 70.3  & 72.8 \\
{\large{\ding{174}}} & {\large{\ding{173}}} -  $D_{s_2}$ ($D_{s_1}$) & 79.9 & 67.1 & 65.8 & 68.4  & 70.3            
    \\
\bottomrule
\end{tabular}
}
\caption{Ablation study of our proposed method. \ourmethod{} is fine-tuned on the combination of all generated instruction datasets.}
\label{ablation_study}
\end{table}

\begin{figure}[t]
\begin{center}
	\includegraphics[width=0.9\columnwidth]{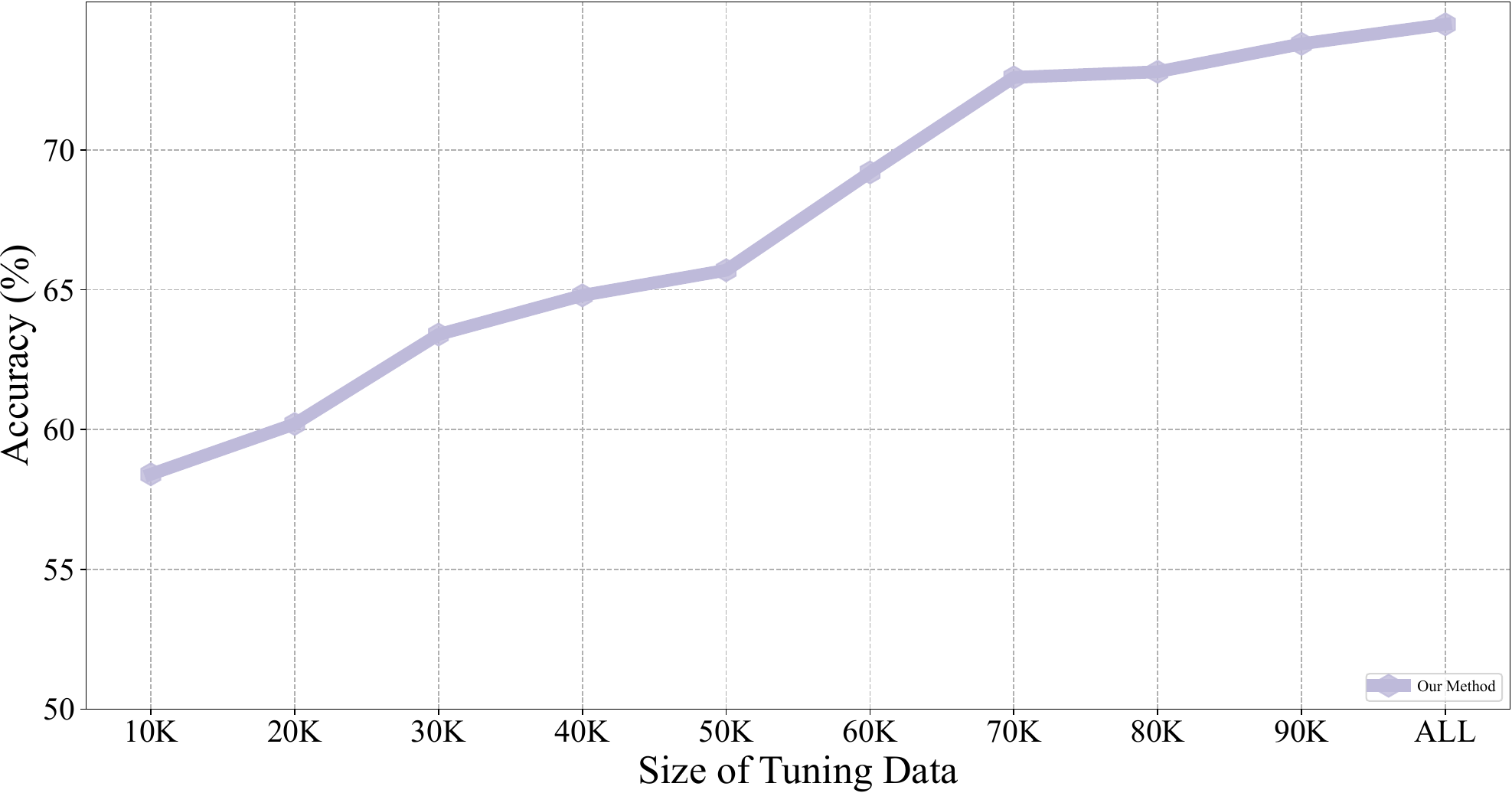}
	\caption{Evaluation results (average scores of 8 programming languages) of Pass@1 on the MultiPL-E with different training sizes by randomly down-sampling.}
	\label{data_size}
\end{center}
\vspace{-15pt}
\end{figure}

\paragraph{Effect of Instruction Data Size.}
To discuss the effect of the size of our created instruction dataset \instruct{} (nearly 97.3K sentences), we plot evaluation scores with different training data sizes in Figure \ref{data_size}. We randomly sample $\{1K, \dots, ALL\}$ sentences from the whole corpora to fine-tune the base Qwen2.5-Coder. With the training data size increasing, the fine-tuned model gets better performance. Surprisingly, only $50K$ pseudo annotated sentences bring large improvement to the multilingual code generation, which benefits from the knowledge transfer of the multilingual agents. 

\begin{figure}[t]
    \centering
    \subfigure[Problem (SFT)]{
    \includegraphics[width=0.45\columnwidth]{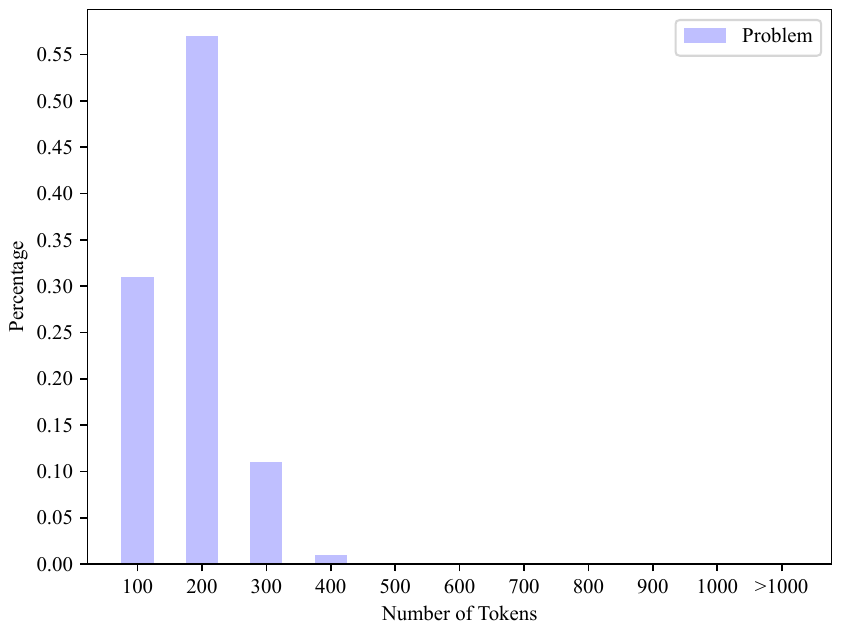}
    \label{count_problem_tokens.pdf}
    }
    \subfigure[Solution (SFT)]{
    \includegraphics[width=0.45\columnwidth]{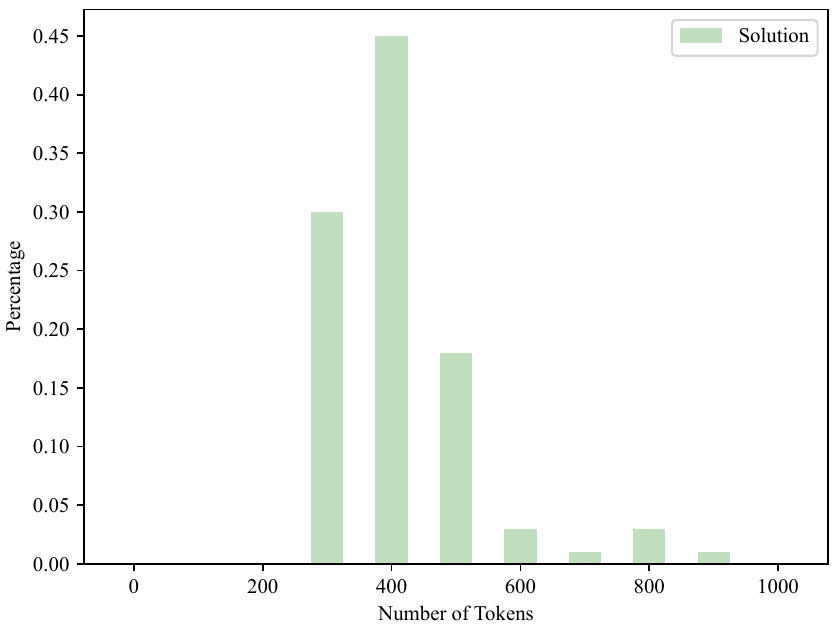}
    \label{count_solution_tokens}
    }
    \subfigure[Problem (DPO)]{
    \includegraphics[width=0.45\columnwidth]{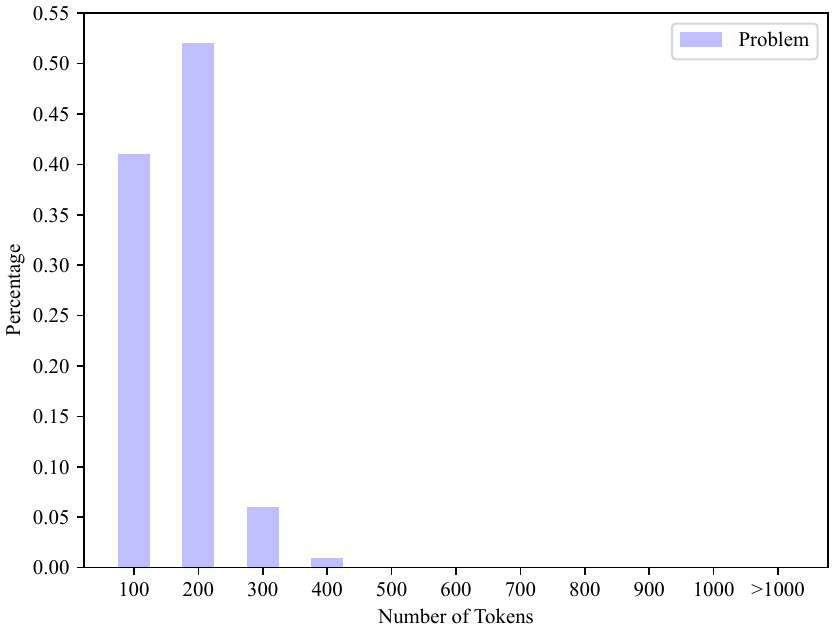}
    \label{count_problem_tokens.pdf}
    }
    \subfigure[Solution (DPO)]{
    \includegraphics[width=0.45\columnwidth]{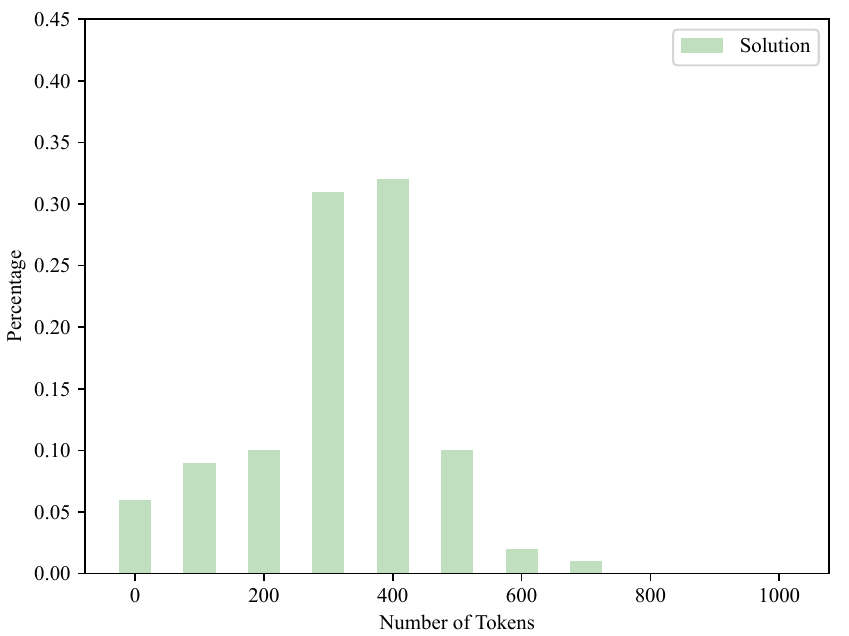}
    \label{count_solution_tokens}
    }
    \caption{Length distribution of problem and solution for the SFT data and DPO data.} 
    \label{token_count}
\end{figure}

\paragraph{Token Count Distribution.}
Figure \ref{token_count} illustrates the distribution of lengths for both the generated problems and their corresponding solutions. On the horizontal axis, we quantify length in terms of the number of tokens comprising each problem or solution. The solution has a similar length distribution to the distribution of the problem. 

\section{Related Work}
\paragraph{Code-related Tasks.}
Pre-training has significantly enhanced the model capabilities of code understanding and synthesis in downstream various tasks, such as CodeBERT \cite{CodeBERT} and CodeT5 \cite{CodeT5}. The model architecture and pre-training objectives originating from natural language processing (NLP) \cite{CodeXGLUE,CodeTransOcean,refine_gpt,chat_unitest} have been increasingly adopted to synthesize programs from human language and perform code infilling, effectively addressing a multitude of software engineering challenges, such as code summarization, code refinement, and code translation.

\paragraph{Code-specific Large Language Model.}
Code-specific large language models (LLMs) \cite{starcoder,code_llama,guo2024deepseekcoder,codearena,execrepobench} trained on large-scale code corpora show remarkable performance across a diverse set of software engineering tasks. Code LLMs culminate in a foundational competence in general code generation and understanding, such as CodeGen \cite{codegen} and Code Llama \cite{code_llama}, which enables them to tackle code-related tasks with better performance. Inspired by the success of multi-agent collaboration in other fields \cite{multi_agents_survey,autonomous_agents_survey}, we introduce the language-specific agent to formulate a multilingual instruction dataset. 

\paragraph{Multilingual Code Instruction Tuning.} 
Instruction tuning is a powerful paradigm enhancing the performance of LLMs by fine-tuning them with the instruction dataset \cite{instructGPT,llama_adapter,self_instructions}. Instruction tuning enables LLMs to generalize better and follow instructions more directly. The previous works \cite{self_instructions,codealpaca} use a foundation LLM to generate the instruction data and then refine the model through instruction tuning with the synthetic data. To further enhance Self-Instruct, WizardCoder \cite{wizardcoder} introduces code Evol-Instruct to produce more high-quality data by using heuristic prompts to increase the complexity and diversity of synthetic data. Recently, OSS-Instruct~\cite{magicoder} and CodeOcean~\cite{wavecoder} leveraged real-world code snippets to inspire LLMs to generate more controllable and realistic instruction corpora. Further, a series of multilingual benchmarks~\cite{multipl_e,mceval,mdeval,fullstack,bigcodebench} (e.g. MultiPl-E, McEval, and MdEval) are proposed to evaluate the multilingual capabilities of code LLMs.

\section{Conclusion}
In this work, we propose a novel multilingual multi-agent collaboration framework to bridge the language divide in programming by generating a comprehensive and high-quality multilingual instruction dataset \instruct{} for fine-tuning \ourmethod{}. Our proposed model leverages the expertise of individual agents, each fluent in a different programming language, to achieve effective knowledge transfer. The collaborative efforts among multiple agents, informed by their individual generation histories, enable the synthesis of versatile programming instructions and solutions. This enriches the instruction dataset and bolsters the capability of our model to generalize across languages, promising significant advancements in the field of multilingual programming development.

\section*{Limitations}
We acknowledge the following limitations of this study: (1) This work focuses on exploring instruction tuning for multilingual code-related works. The investigation of this paradigm on other multilingual tasks has not been studied yet. (2) While our approach aims to facilitate knowledge transfer across multiple programming languages, it may not be equally effective for all languages, potentially leading to a bias towards more commonly used or better-represented languages in the dataset.

\bibliography{custom}
\bibliographystyle{acl_natbib}

\clearpage
\end{CJK*}
\end{document}